\documentclass[conference]{IEEEtran}
\IEEEoverridecommandlockouts
\usepackage[T1]{fontenc}
\usepackage{cite}
\usepackage{amsmath,amssymb,amsfonts}
\usepackage{algorithm}
\usepackage{graphicx}
\usepackage{epstopdf}
\usepackage{textcomp}
\usepackage{xcolor}
\usepackage{algpseudocode}
\def\BibTeX{{\rm B\kern-.05em{\sc i\kern-.025em b}\kern-.08em
    T\kern-.1667em\lower.7ex\hbox{E}\kern-.125emX}}
    
\DeclareMathOperator{\EX}{\mathbb{E}}
    
\begin{document}

\title{Dense Random Texture Detection using Beta Distribution Statistics \\
\thanks{Thanks to Panasonic Automotive Systems for funding this research.}
}

\author{\IEEEauthorblockN{S\"oren Molander}
\IEEEauthorblockA{\textit{ADAS Department} \\
\textit{Panasonic Automotive Systems Europe}\\
Langen, Germany \\
soeren.molander@eu.panasonic.com}
}

\maketitle

\begin{abstract}
This note describes a method for detecting dense random texture using fully connected points sampled on image edges. An edge image is randomly sampled with points, the standard L2 distance is calculated between all connected points in a neighbourhood. For each point, a check is made if the point intersects with an image edge. If this is the case, a unity value is added to the distance, otherwise zero. From this an edge excess index is calculated for the fully connected edge graph in the range [1.0..2.0], where 1.0 indicate no edges. The ratio can be interpreted as a sampled Bernoulli process with unknown probability. 
The Bayesian posterior estimate of the probability can be associated with its conjugate prior which is a Beta($\alpha,\beta$) distribution, with hyper parameters $\alpha$ and $\beta$ related to the number of edge crossings. Low values of $\beta$ indicate a texture rich area, higher values less rich. The method has been applied to real-time SLAM-based moving object detection, where points are confined to tracked boxes (rois). 

\end{abstract}


\section{Introduction}
Texture detection is one of the most established fields in computer vision. It is therefore a valid question to ask if yet another method is needed to add to an already well established field. Many methods, however, are not well suited for real-time applications, or have other issues. Real-time detection of moving objects is such an example, which is an active area of research with applications in e.g. ADAS for the automotive industry. For one particular application developed, 2D points are classified into dynamic or non-dynamic points using methods involving SLAM described elsewhere (\cite{Golla2023}). The point clouds are clustered into boxes which are used in a tracking system. In some cases, this box formation process fails in areas with dense random texture and a method was therefore needed to deal with this problem. 
 
\subsection{Related work} 
A survey of statistical methods is found in \cite{Ramola2020}, which classifies texture detection into three categories:

\begin{itemize}
\item Structural approaches (e.g. morphology and edges)
\item Model-based approaches (e.g. autoregessive models and fractals)
\item Transformation-based approaches (e.g. wavelet and Gabor transforms)
\end{itemize}

The present methods falls naturally in the structural approach. 
Examples of texture detection and classification include co-occurrance matrices \cite{haralick73texture}, local binary patterns \cite{DongchenHe1990} and Gabor wavelets \cite{Huo2019}. Beta distributions have been used in e.g. texture modelling for SAR images \cite{Arai521720} and in modelling medical phantom images \cite{Chui2013} and reflectance modelling \cite{ATTEWELL2007548}.

\subsection{Motivation} \label{motivation}
Most texture detection methods focus on finding and classifying textures. In the particular application here, the information of interest was to detect the density of random textures and to discard high-texture areas. This is a binary decision problem, and no attempt was made to classify the exact type of texture from a library.

\subsection{Main Contributions}
The main contribution in this paper is a straightforward method to classify textures into two classes: $\{High\ random\ texture, Low\ random\ texture\}$. The method is light-weight and is easily implemented in embedded systems.

\section{Overview of method}
The method comprises the following steps:
\begin{itemize}
\item Calculate a gradient image
\item Sample points on gradient edges in a region of interest
\item For each pair of points, connect to all other points and calculate the L2 distance $d$. 
\item For each pair of points $\{p_i$,$p_j\}$, add unity if the point crosses a gradient > $T_{grad},$ and call it "path excess". Repeat for all points across the edge from $p_i$ and $p_j$ and calculate the sum $w=\sum\limits_{k}\textbf[1_k]_A$,
where A = $\{crosses\ edge\}$ and the sum is over all pixels $\{k\}$ from $p_i$ to $p_j$, and $\textbf[1_k]_A$ the indicator function. 
\item Calculate the excess index along a graph edge $E=(\frac{2\times w}{w+d})$, $d=||p_i - p_j||$, and $E\in [1,2]$.
\item Repeat for all $\frac{n(n-1)}{2}$ pairs $\{p_i,p_j\}$ in the point set.
\item The total length of the graph = $L$ and total edge excess=$E_{L}\times L$
\item Define the total graph edge excess as $PE = (1+\frac{E_L}{L})$
\end{itemize}

\begin{figure}[h!]
  \centering
  \includegraphics[width=.40\textwidth]{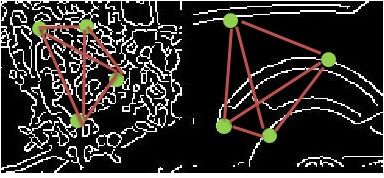}\hfill
  \includegraphics[width=.40\textwidth]{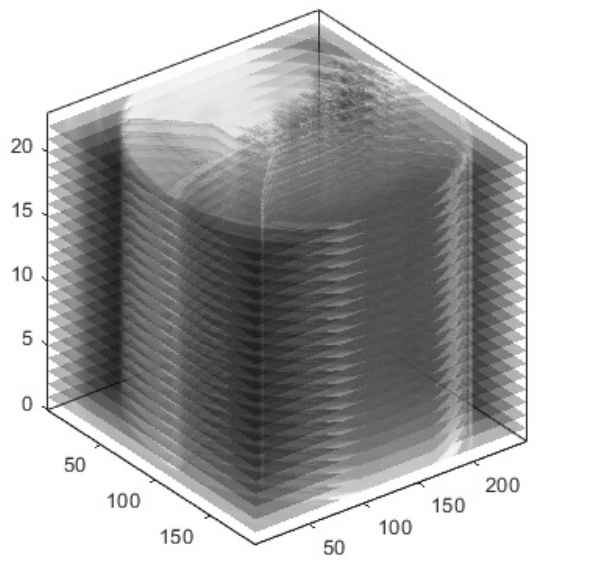}
  \caption{The general idea of an edge graph. \emph{Top Left}: High density texture, with many crossings. \emph{Top Right}: Low density graph. \emph{ Bottom]}: The edge excess distributions have been derived over boxes in a sequence of images.}
\label{fig:lowTextureHist}
\end{figure}

\begin{figure}[h!]
  \centering
  \includegraphics[width=.22\textwidth]{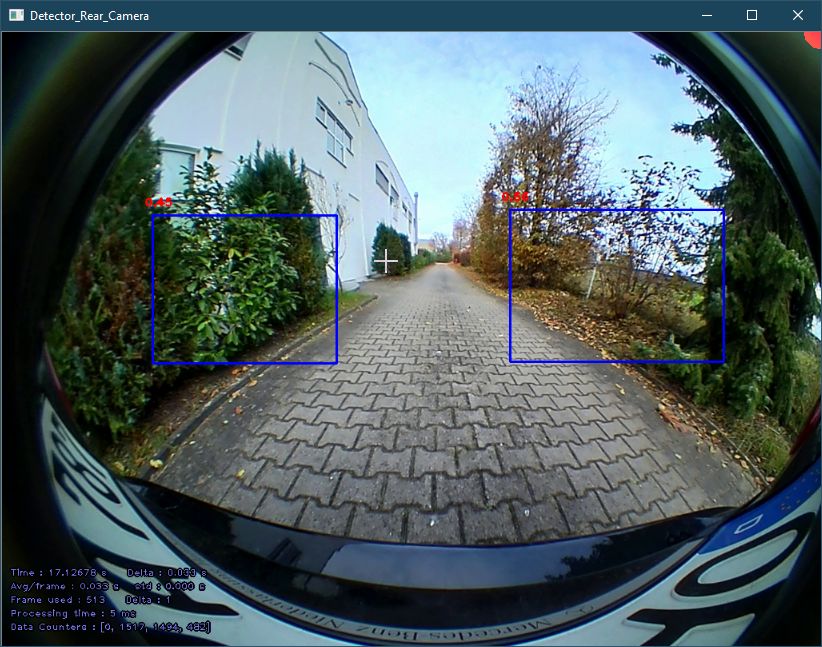}\hfill
  \includegraphics[width=.22\textwidth]{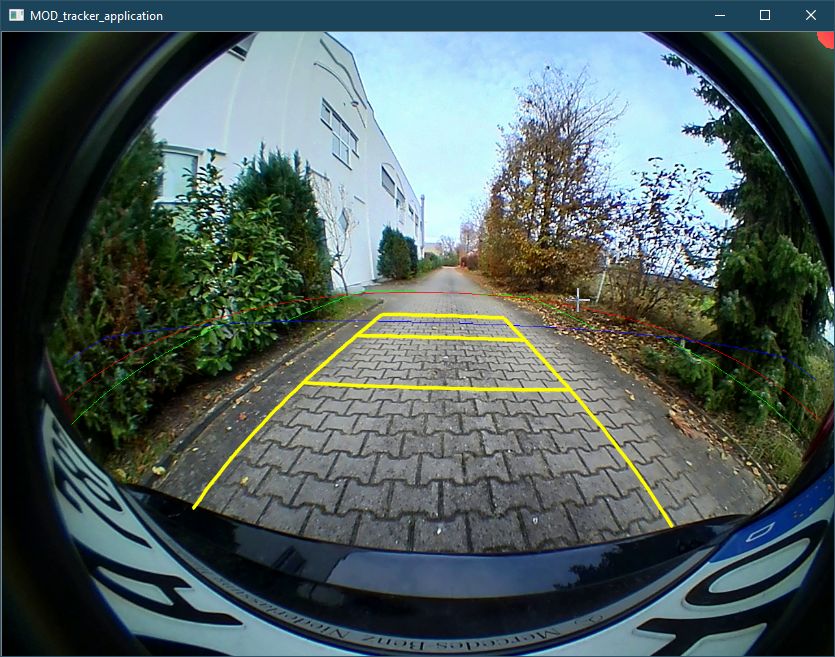}
  \caption{Illustration of false detections but no false tracks.       \emph{Left}: Falsely detected rois/boxes, used in generating edge excess statistics. \emph{Right}: No false tracks.}
\label{fig:falseDetect}
\end{figure}

\begin{figure}[!h]
  \centering
  \includegraphics[width=.50\textwidth]{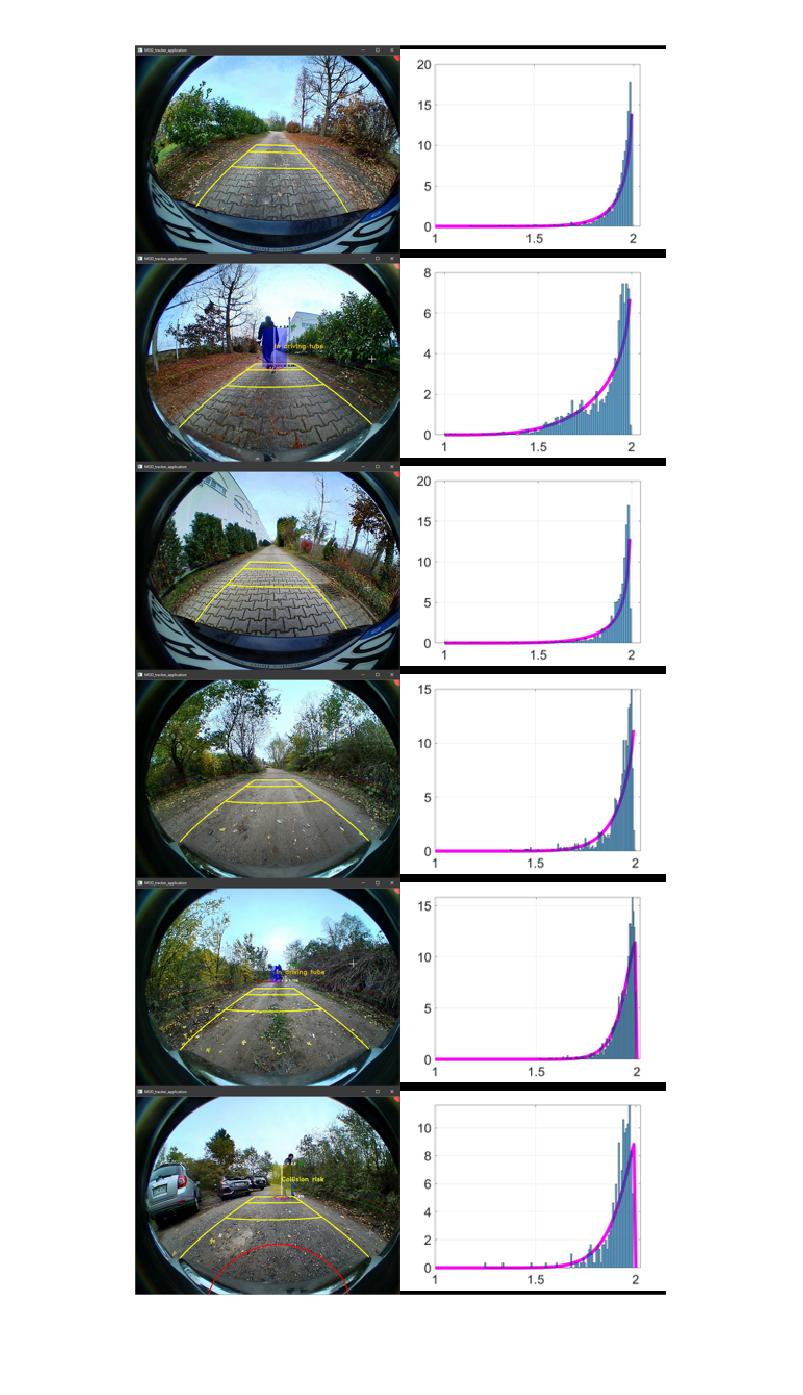}
  \caption{\textit{Left column:} Images with a high degree of random texture. \textit{Right column:} A histogram of the edge excess index with the fitted Beta distribution. Edge excess statistics is retained for $all$ rois, even those that are not retained in the tracking filter.}
\label{fig:highTextureHist}
\end{figure}

Given a region of interest (roi), a point set sampled on edges, the procedure is summarized in \textbf{Algorithm} \ref{alg:TextureDetection}. In the application here the sampling was performed over rois from a sequence of images, using a standard roi/box tracking system (using an extended Kalman filter), and where the box foot mid-point touching the ground was used in the tracking filter. In order to form a track, N detections (typically $N=5$) are needed. This means that if tracking boxes/rois exist for shorter number of frames < 5, they are never formed into a track. Thus, the retained tracking boxes shown in figure \ref{fig:highTextureHist} show $low\ texture$ boxes, but the statistics have been calculated for $all$ boxes (not shown as a track). This is
illustrated in figure \ref{fig:falseDetect}.

\begin{algorithm}[H]
\caption{Texture Calculation}
\label{alg:TextureDetection}
\begin{algorithmic}[1]
  \Procedure{Texture}{$Image, points[],Roi[],T_{grad}$}
  \For{\texttt{r in $Roi[]$}}
     \State $gradIm \gets getEdgeImage(Image,T_{grad}, Roi[r])$
     \State $getPathLengths(gradIm,points,L,E_L)$
     \State $pathExcess \gets (1+\frac{E_L}{L})$
  \EndFor
  \EndProcedure
\end{algorithmic}
\end{algorithm}

\begin{algorithm}[H]
\caption{High Texture Rejection}
\label{alg:TextureRejection}
\begin{algorithmic}[1]
  \Procedure{Texture}{$pathExcess,Roi[],T_{pe},i$}
  \For{\texttt{r in $Roi[]$}}
     \State $keepRoi \gets \EX_i(pathExcess) < T_{pe}$
  \EndFor
  \EndProcedure
\end{algorithmic}
\end{algorithm}

The edge counting method can be seen as a set of Bernoulli trials:

\[
\left\{
\begin{array}{c}
  Pr(crossed\ edge) = p  \\
  Pr(no\ crossed\ edge) = 1-p
\end{array}
\right.
\]

Since the exact probability $p$ is not known, it can be estimated using the Bayesian posterior Beta distribution with hyper parameters $\alpha$ and $\beta$ \cite{Johnson1995} (for $\alpha=1$ and $\beta=1$ it is uniform distribution). 

\begin{equation}
Pr(p|\alpha,\beta) = \frac{p^{\alpha-1}(1-p)^{\beta-1}}{B(\alpha,\beta)},\\
B(\alpha,\beta)=\frac{\Gamma(\alpha)\Gamma(\beta)}{\Gamma(\alpha+\beta)}
\end{equation}

The edge excess histograms follow the Beta distribution well. For high-texture images the beta parameter tend to be low, for low-texture images higher. 
This is indicated in figures \ref{fig:lowTextureHist} and \ref{fig:highTextureHist}. The edge excess histograms were calculated using a slam-based detection and tracking system. Detected boxes include both tracking of moving objects (shown in the pictures), as well as falsely detected boxes, which are deleted and not tracked. The Beta distribution fits were calculated by estimating the mean and the variance over entire sequences:

\begin{equation}
\begin{array}{l}
\alpha= (\frac{\mu(1-\mu)}{var}-\frac{1}{\mu}) \times \mu^2 \\
\beta=\alpha \times (\frac{1}{\mu}-1)
\end{array}
\end{equation}

\begin{figure}[!h]
  \centering
  \includegraphics[width=.50\textwidth]{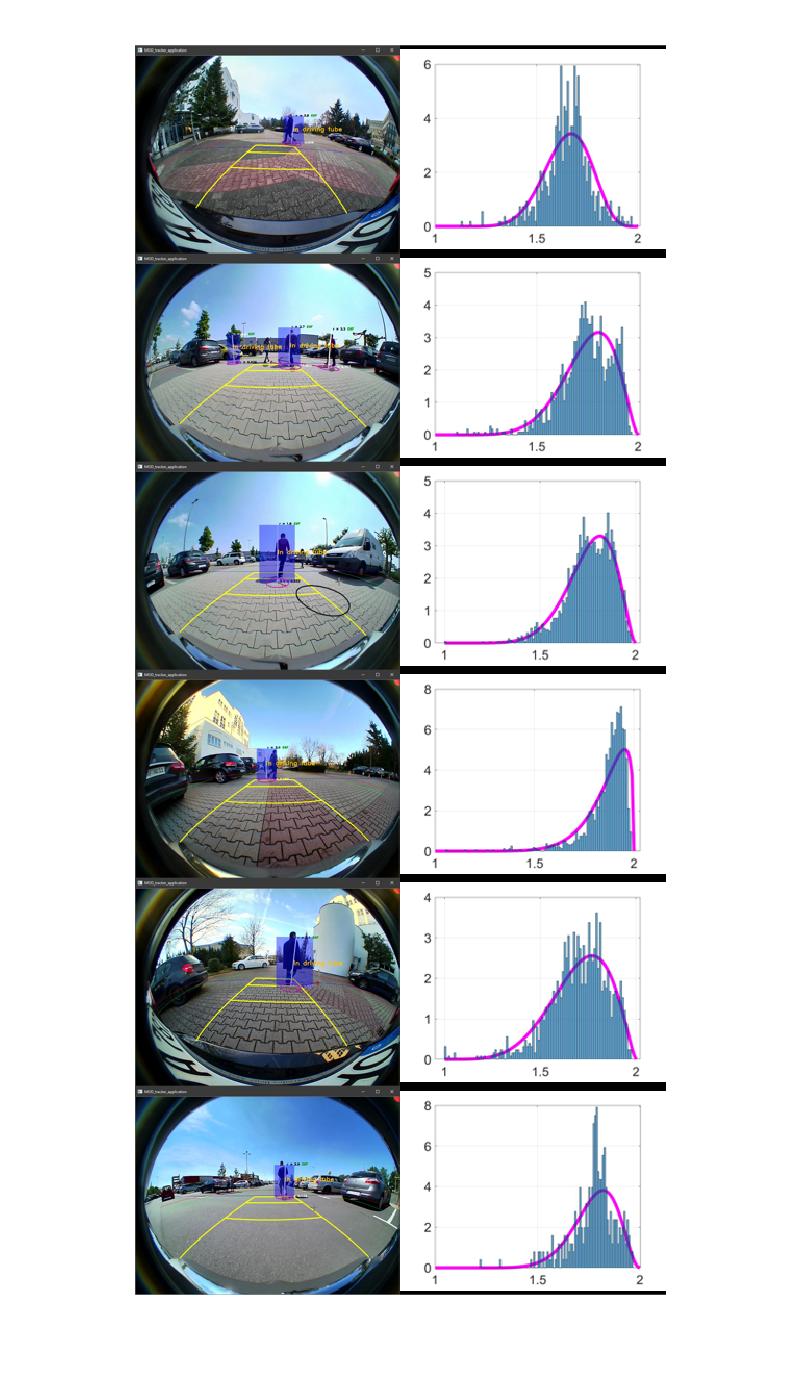}
  \caption{\textit{Left column:} Low texture images containing moving objects.  Blue boxes indicate tracked moving objects in the driving tube. \textit{Right column:} A histogram of the edge excess index with the fitted Beta distribution. The standard Beta distribution has been shifted from the standard interval [0,1] to [1,2]. }
\label{fig:lowTextureHist}
\end{figure}

\begin{figure}[H]
  \centering
  \includegraphics[width=.50\textwidth]{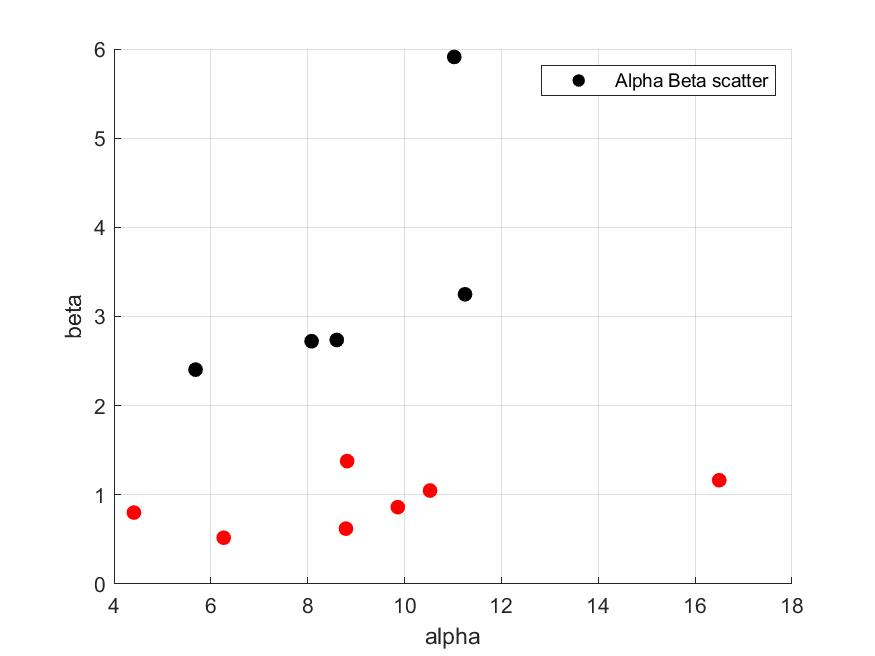}
  \caption{A scatter plot of the hyper parameters in figures \ref{fig:lowTextureHist} and \ref{fig:highTextureHist}.High-texture Beta values < 1.5 are marked in red.}
\label{fig:paramClusters}
\end{figure}

In order to extract threshold for classification of low and high texture, a scatter plot was made threshold was developed for the data set at hand (figure \ref{fig:paramClusters}). High-texture regions are typically characterized as sharply peaked low-beta and high alpha values (figure \ref{fig:highTextureHist}). Low texture, conversely have flatter high-alpha distributions, typically with a maximum peak (figure \ref{fig:lowTextureHist}). Because the distributions were generated for a time-sequence of images with detections in tracking boxes, the distribution parameters can vary and sometimes be a mixtures of distributions.

In order to develop a practical method for removing ROIs with high-density textures, a recursive average of the edge excess was calculated over ROIs in sequences. In high-texture areas, eventually the average will fall over a selected threshold and false detections over textures will be removed. Conversely, after transition period in low texture areas, the edge excess will converge to a value below the threshold. The actual threshold varies depending on the data and camera, for the data set shown here a value of $T_{pe}=1.9$ was chosen. This is indicated in \textbf{Algorithm} \ref{alg:TextureRejection}, where the expection value $\EX_i(.)$ is a standard recursive average over the sequence for the current image frame $i$.

\section{Conclusion}
A light-weight method for detection of high random texture has been developed and tested on real-world sequences. Future work will include adaptive threshold mechanisms using on line learning algorithms. 

\bibliographystyle{ieee}
\bibliography{biblio}

\end{document}